\pdfoutput=1
\documentclass[11pt]{article}

\usepackage[]{naacl2024}
\usepackage{times}
\usepackage{latexsym}
\usepackage{graphicx}
\usepackage{tabularx}
\usepackage{multirow}
\usepackage{amsmath} 
\usepackage{comment}
\usepackage{booktabs}
\usepackage{stix,bbding,pifont,utfsym,fontawesome}

\usepackage[T1]{fontenc}
\usepackage[utf8]{inputenc}
\usepackage{soul}
\usepackage{microtype}
\title{The Role of Chain-of-Thought in Complex Vision-Language Reasoning Task}

\author{Yifan Wu\textsuperscript{1}, Pengchuan Zhang\textsuperscript{2}, Wenhan Xiong\textsuperscript{2}, Barlas Oguz\textsuperscript{2}, James C. Gee\textsuperscript{1}, Yixin Nie\textsuperscript{2} \\
\textsuperscript{1}University of Pennsylvania \quad \textsuperscript{2}Meta AI\\
yfwu@seas.upenn.edu \quad gee@upenn.edu \\
\{pengchuanzhang, xwhan, barlaso, ynie\}@meta.com}

\begin{document}
\pagestyle{plain}
\maketitle
\begin{abstract}
The study explores the effectiveness of the Chain-of-Thought approach, known for its proficiency in language tasks by breaking them down into sub-tasks and intermediate steps, in improving vision-language tasks that demand sophisticated perception and reasoning. We present the "Description then Decision" strategy, which is inspired by how humans process signals. This strategy significantly improves probing task performance by 50\%, establishing the groundwork for future research on reasoning paradigms in complex vision-language tasks.
\end{abstract}

\section{Introduction}
Large language models (LLMs) have shown impressive performance in many language tasks, fostering the development of general AI assistants. An emerging trend in AI research aims to expand the potential of LLMs beyond text perception, incorporating visual data for more comprehensive models \cite{zhu2023minigpt, liu2023llava, chen2023minigptv2}. GPT-4V(ision) was recently released and has garnered significant interest for its exceptional abilities in multimodal perception and reasoning. 

However, in complex vision-language tasks, although GPT-4V significantly outperforms the current state-of-the-art, it still lags behind human performance, as demonstrated in Section~\ref{sec:eval}. We try to understand this challenge. Visual understanding extends beyond mere perception \cite{zellers2019recognition}. Complex visual-language tasks demand recognition-level perception, such as localizing and classifying objects and their attributes, as well as cognition-level reasoning, such as inferring intents, goals, and temporal and social dynamics. Humans can seamlessly integrate these two stages, but this remains challenging for 
LLMs with vision modality (vision-LLMs). 

The Chain-of-Thought strategy, known for its effectiveness in language tasks by breaking them into sub-tasks with intermediate steps, has been studied extensively \cite{wei2022chain, yao2023tree, lyu2023faithful}, In this research, we investigate whether this method can enhance vision-language tasks, particularly those requiring complex reasoning. However, when confronting complex visiolinguistic tasks, current LLMs struggle to figure out the proper reasoning paradigm by themselves. Unlike math problems, which have a specific task-driven reasoning approach, these tasks may not have a clear and unique reasoning path leading to the final output. 

\begin{figure}[t]
\includegraphics[width=\linewidth]{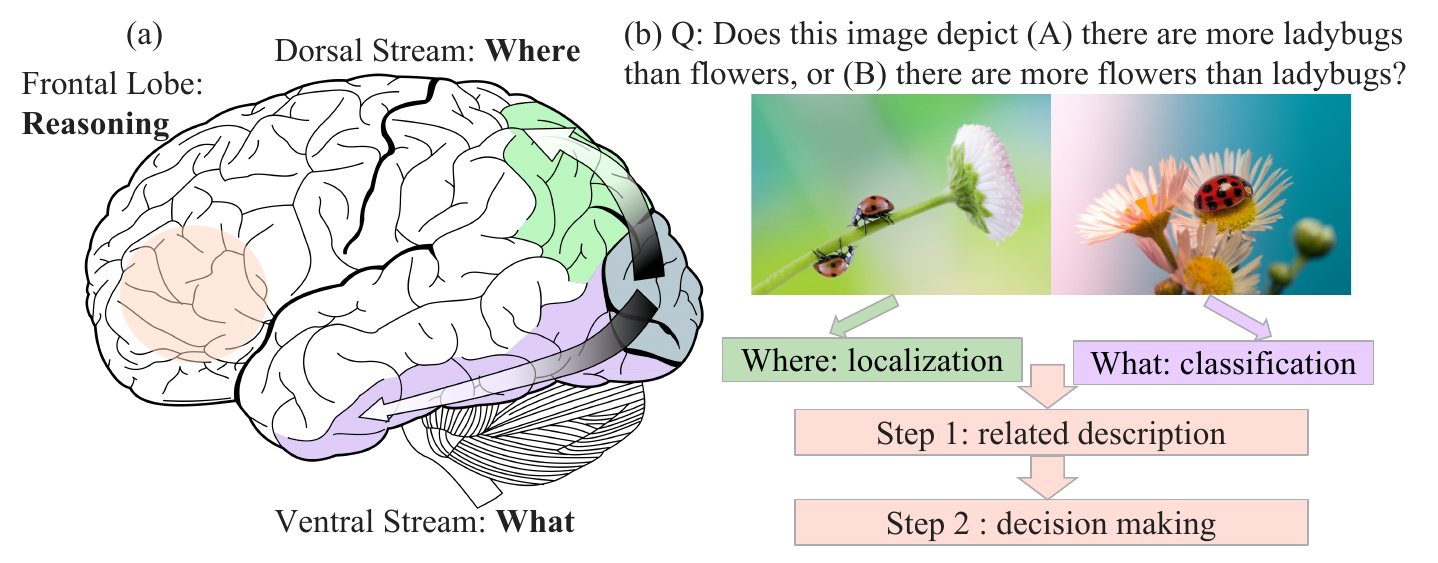}
\caption{\textbf{Brain-inspired Two-step Reasoning.} Figure (a) is adapted from Wikipedia under license CC BY-SA 3.0 DEED. Figure (b) is adapted from Winoground \cite{thrush2022winoground}.}
\label{fig: schematic}
\end{figure}

\begin{figure*}[t]
\centering
\includegraphics[width=\linewidth]{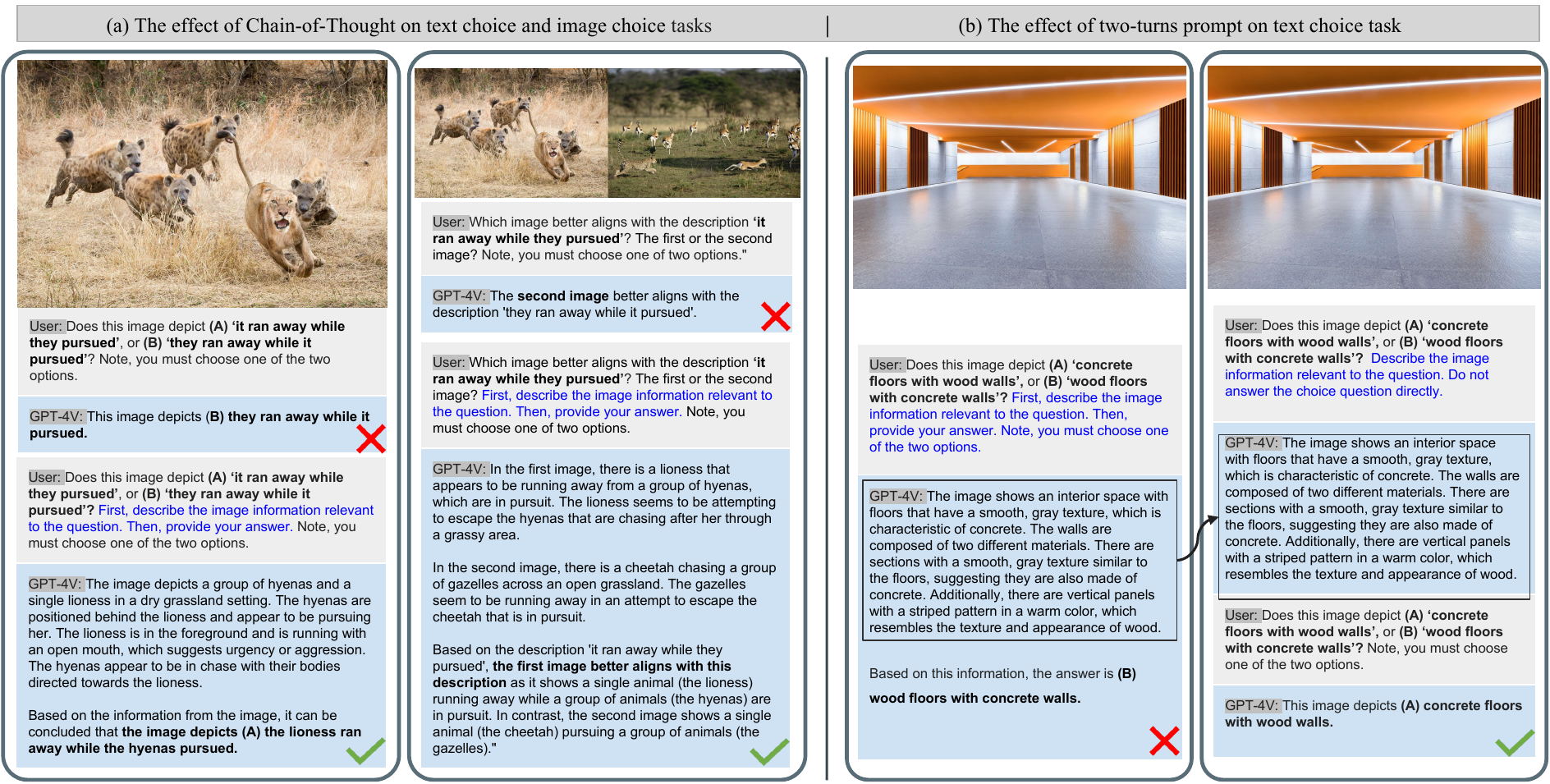}
\caption{Examples of results with different prompt configurations. Text in blue highlights differences in the prompts. All figures shown here are from Winoground \cite{thrush2022winoground}}
\label{fig: method}
\end{figure*}

Research on human cognition provides clues to a proper reasoning mode. Visual information propagates through two streams (Figure~\ref{fig: schematic} (a)). The ventral stream (or the `what pathway') is involved with object identification, while the dorsal stream (or the `where pathway') processes objects' spatial location. These two streams decouple recognition into local processing modules. The cognition part, i.e., the reasoning and decision-making function, is mainly executed in the frontal lobe. Such modularization is a significant characteristic of brain structure and function \cite{gu2015emergence, bassett2017network}. Inspired by how humans process signals, we design prompts with specialized modules for steps with different emphases.  

The latest initial attempt at Chain-of-Thought on vision-language tasks focuses on the recognition task \cite{yang2023set}, while another only provides qualitative analysis \cite{yang2023dawn}. In this work, we analyze the prompting strategy for complex vision-language tasks analogous to the brain's information processing. Our \emph{Description (information-extracting) then Decision (decision-making)} strategy consistently improves the performance across different experimental settings.

\section{Probing Task}
\label{sec: task}
The ideal probing task should present sufficient challenges in both visual recognition and text comprehension. Furthermore, it should demand intricate reasoning, implying that there is a logical connection between the image and text facts leading to the final output. Consequently, we selected Winoground \cite{thrush2022winoground} as a case study for our experiments.
Winoground is both a dataset and a task designed specifically to evaluate visio-linguistic compositional reasoning. This task involves being given two images and two captions, with the goal of correctly matching each image with its corresponding caption. Notably, both captions utilize the exact same set of words, but they are arranged in different orders so that each caption primarily describes one of the two images.

The Winoground task is notably challenging \cite{diwan2022winoground}. It requires not only a robust visual recognition ability to identify small or blurred objects and differentiate attributes and actions among proximate objects but also sophisticated visio-linguistic compositional reasoning. For instance, solving Winoground sometimes necessitates interpreting images non-literally due to the idiomatic language usage in a caption (e.g., “it starts with Z and ends with A” might describe an image of a zebra). In other instances, one might need to infer past or future events from the scenes depicted in the image (e.g., "the cup on the left was filled first, and the cup on the right was filled second"). Examples of Winoground can be found in Fig \ref{fig: method}.

The original Winoground task consists of two experimental setups: the text score and the image score. The text score evaluates the model's ability to select the correct caption from two given captions when provided an image. Conversely, the image score assesses the model's ability to choose the appropriate image from two available options given a caption. The original Winoground was tested based on the feature embedding similarities between captions and images in vision-language models, such as CLIP \cite{radford2021learning}. To assess recent large vision-language models like GPT-4V, we reformulate the Winoground as a choice-based visual question-answering task. Formally, given images $I_0$ and $I_1$ and captions $C_0$ and $C_1$, the text score for a data point ($C_0$, $I_0$, $C_1$, $I_1$) is computed as follows: 
\begin{equation}
\small
s\left(C_0, I_0, C_1, I_1\right)=\left\{\begin{array}{cl}
1 & \text { if } f\left(C_0, C_1, I_0\right) = C_0,\\
& \text { and } f\left(C_0, C_1, I_1\right)=C_1 \\
0 & \text { otherwise }
\end{array}\right.
\label{eqn1}
\end{equation},
where $f(\cdot)$ is the large language model that provides answers through a generation process. For a data point to be classified as correct, both images in a pair must align with their textual descriptions.

Similarly, for the image choice task, the score is determined as follows:
\begin{equation}
\small
s\left(C_0, I_0, C_1, I_1\right)=\left\{\begin{array}{cl}
1 & \text { if } f\left(I_0, I_1, C_0\right) = I_0,\\
& \text { and } f\left(I_0, I_1, C_1\right)=I_1 \\
0 & \text { otherwise }
\end{array}\right.
\label{eqn2}
\end{equation}
The Winoground dataset comprises 400 pairs of images and their corresponding captions.


\section{Evaluation}
\label{sec:eval}
We first present the Winoground benchmark in Table~\ref{benchmark} to evaluate the performance of GPT-4V in comparison to others. Note that while the original task evaluation is based on the encoding similarities between images and text, the setups for large vision-language models may vary slightly among methods. For example, TIFA \cite{hu2023tifa} and VQ2 \cite{yarom2023you} ask a series of questions given one image and one caption, then accumulate the scores. MMICL \cite{zhao2023mmicl} is given two images and two captions in each prompt. While in our set-up, we are given one image with two captions, or two images with one caption, as formulated in Eqn~\ref{eqn1} and Eqn~\ref{eqn2}. However, these variations will not impact the primary objective of our study, which is to analyze the effects of various prompt configurations.

\subsection{The Role of Chain-of-Thought}
We analyze whether using a chain-of-thought prompt strategy, by decomposing the visual-language complex reasoning task into recognition and reasoning steps, can be beneficial compared to directly asking the model for the answer. The quantitative results are reported in Table~\ref{benchmark}, and the qualitative examples are shown in Figure~\ref{fig: method} (a), with the prompt configurations detailed accordingly.In the following, we present the prompt we use to evaluate GPT-4V's text score (\emph{Text}) and image score (\emph{Image}), without and with CoT. 

\noindent\textbf{GPT-4V (Text)}: {[`image-0' or `image-1']} Does this image present (A) {[`caption-0']}, or (B) {[`caption-1']}? Note, you must choose one of the two options.

\noindent\textbf{GPT-4V CoT (Text)}: {[`image-0' or `image-1']} Does this image present (A) {[`caption-0']}, or (B) {[`caption-1']}? First, describe the image information relevant to the question. Then, provide your answer. Note you must choose one of the two options.

\noindent\textbf{GPT-4V (Image)}: {[`image-0']}, {[`image-1']} Which image better aligns with the description {[`caption-0' or `caption-1']}? The first image or the second image? Note you must choose one of two options.

\noindent\textbf{GPT-4V CoT (Image)}: {[`image-0']}, {[`image-1']} Which image better aligns with the description {[`caption-0' or `caption-1']}? The first image or the second image? First, describe the image information relevant to the question. Then, provide your answer. Note you must choose one of two options.

As shown in Table~\ref{benchmark}, there are consistent improvements in both the \emph{Text} and \emph{Image} score settings, leading to a $50\%$ improvement (from 39.25 to 58.75) in the \emph{Group} score. The latter represents the percentage of data points that have both \emph{Text} and \emph{Image} probes answered correctly. The ``Description then Decision" strategy is particularly beneficial for the \emph{Image} score setting, which improves from 46.25 to 68.75. 
We observe that for GPT-4V, the image score is significantly lower than the text score. This could be attributed to the use of multiple images as input. The Chain of Thought (CoT) strategy significantly improves the image score and largely closes the gap with the text score.

\begin{table}
    \small
    \centering
    \begin{tabular}[\linewidth]{lccc}
       \toprule
        \textbf{Model \& Prompt} & \textbf{Text} & \textbf{Image} & \textbf{Group} \\
        \midrule
        Random Chance & 25.00 & 25.00 & 16.67 \\
        MTurk Human & 89.50 & 88.50 & 85.50 \\
        \midrule
        \multicolumn{4}{c}{\textbf{CLIP-based encoding similarity}} \\
        \midrule
        CLIP \cite{radford2021learning}  & 30.75 & 10.50 & 8.00 \\
        METER \cite{dou2022empirical}    & 44.99 & 22.75 & 18.75 \\
        Fiber \cite{wang2023equivariant} & 51.49 & 31.49 & 27.50  \\
        \midrule
        \multicolumn{4}{c}{\textbf{Vision Large Language Models}} \\
        \midrule
        TIFA \cite{hu2023tifa}  & 19.00 & 12.50 & 11.30 \\
        PALI \cite{chen2022pali}& 46.50 & 38.00 & 28.75 \\
        VQ2 \cite{yarom2023you} & 47.00 & 42.20 & 30.50 \\
        MMICL \cite{zhao2023mmicl} & 45.50 & 44.99 & 43.00 \\
        GPT-4V & 69.25 & 46.25 & 39.25 \\
        GPT-4V CoT & \textbf{75.25} & \textbf{68.75} & \textbf{58.75} \\
       \bottomrule
    \end{tabular}
    \caption{\textbf{Results on the Winoground task.} The  scores (\%) for various models are reported in the following references: MTurk Human and CLIP in \cite{thrush2022winoground}, METER and Fiber in \cite{wang2023equivariant}, TIFA, 
    PALI, and VQ2 in \cite{yarom2023you}, and MMICL in \cite{zhao2023mmicl}. Note that the CLIP-based encoding similarity scores are derived from a deterministic process, while the results for the large vision-language models are obtained through a generative process.}
\label{benchmark}
\end{table}

The difference in generative processes between these two prompts is as follows: for GPT-4V, the answer is conditioned on the image and question:
\begin{equation}
\small
\text{P(Answer | Image, Question)}
\end{equation}
While for GPT-4V CoT, the answer is generated based on the image, question, and description:
\begin{equation}
\small
\text{P(Answer | Image, Question, Description)}
\end{equation}

Although generating descriptions from images does not introduce new information, our results show that this step simplifies the reasoning or decision-making for the model by translating visual signals into textual ones. To assess generalizability, we conducted the same experiments on other vision-based large language models, LLaVA \cite{liu2023llava} and InstructBLIP \cite{dai2023instructblip}, as shown in Table~\ref{table:ablation} (a). The ``description then decision" strategy consistently improves performance.

\begin{figure*}[t]
\centering
\includegraphics[width=0.9\linewidth]{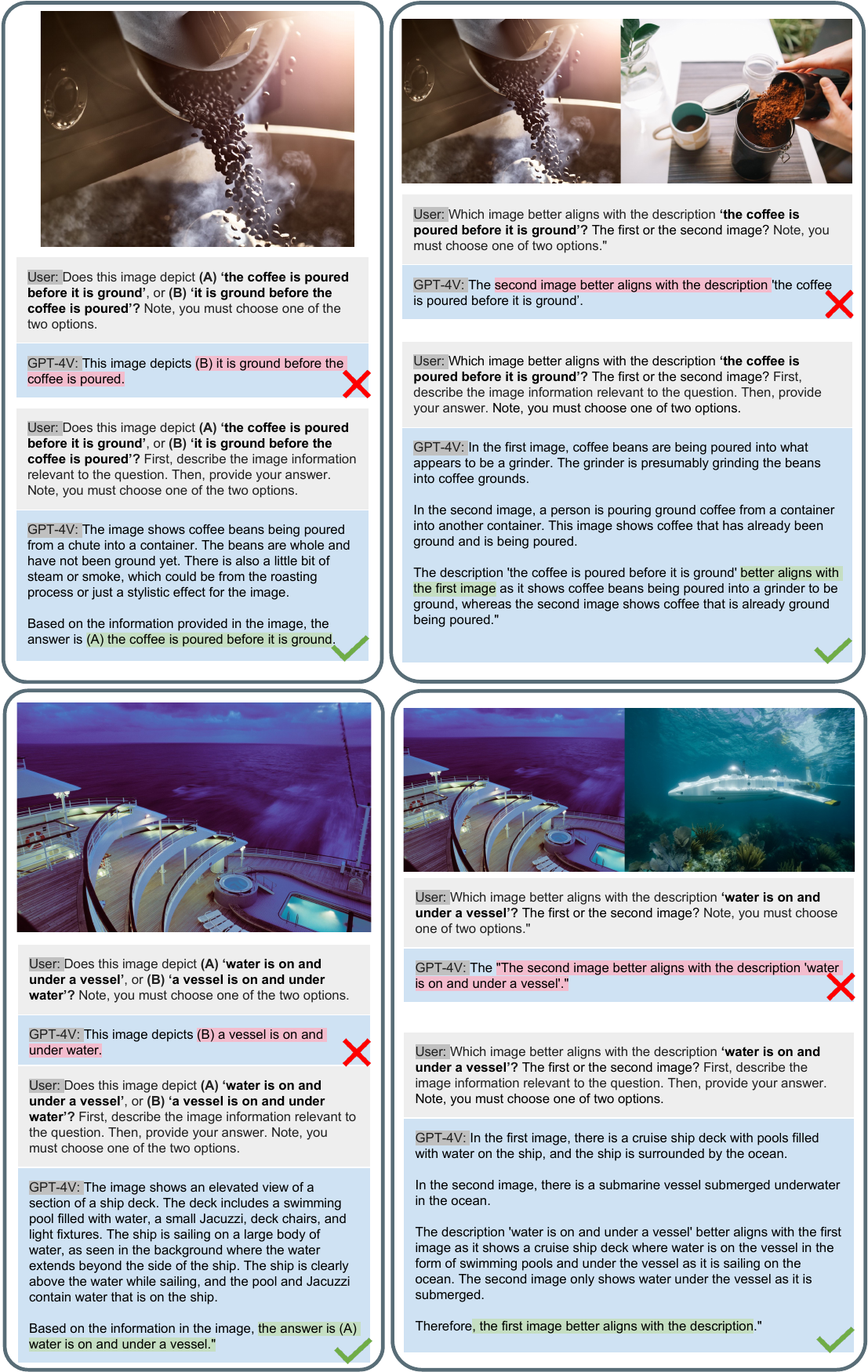}
\caption{The qualitative examples demonstrate the effect of Chain-of-Thought. For each example, the left side illustrates the \emph{text} choice setting and the right side depicts the \emph{image} choice setting. The top portion shows the outcome without Chain-of-Thought,  referred to as "GPT-4V (1-turn)," while the bottom part shows the results with Chain-of-Thought, labeled as "GPT-4V CoT (1-turn)" in Table~\ref{benchmark}. All images shown here are from Winoground \cite{thrush2022winoground}.}
\label{fig: result1}
\end{figure*}

\begin{figure*}[t]
\centering
\includegraphics[width=0.9\linewidth]{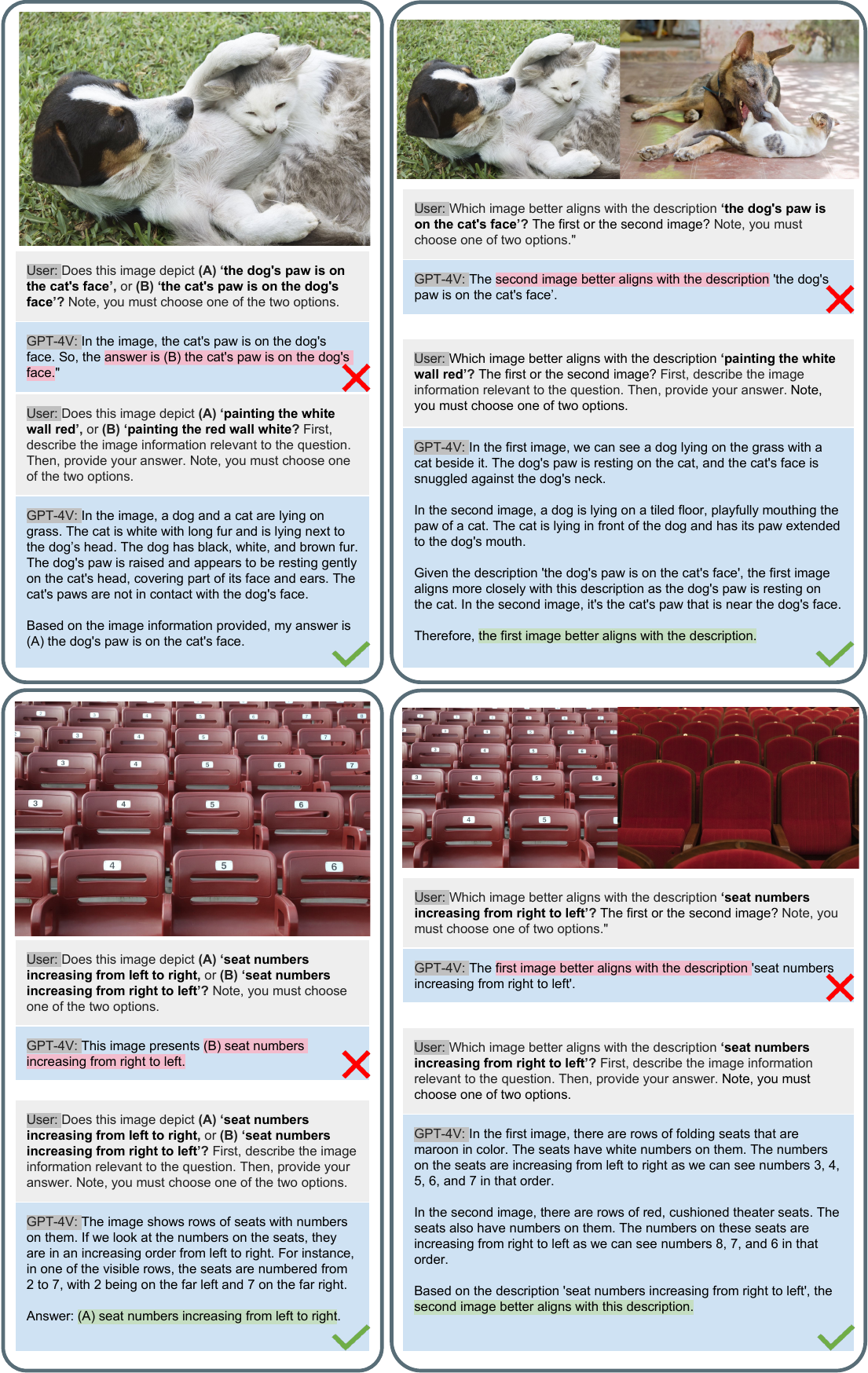}
\caption{The qualitative examples (continued) demonstrate the effect of Chain-of-Thought. For each example, the left side illustrates the \emph{text} choice setting and the right side depicts the \emph{image} choice setting. The top portion shows the outcome without Chain-of-Thought,  referred to as "GPT-4V (1-turn)," while the bottom part shows the results with Chain-of-Thought, labeled as "GPT-4V CoT (1-turn)" in Table~\ref{benchmark}. All images shown here are from Winoground \cite{thrush2022winoground}.}
\label{fig: result2}
\end{figure*}

More qualitative results demonstrating the effectiveness of our ``Description then Decision'' prompt strategy 
are shown in Figure~\ref{fig: result1} and Figure~\ref{fig: result2}.

\begin{table}[t]
\small
\centering
\setlength\tabcolsep{12pt}
\begin{tabular}[\linewidth]{lr}
\toprule
\textbf{Model \& Prompt} & \textbf{Text}  \\
\midrule
\multicolumn{2}{c}{\textbf{(a) The Effect of CoT on Other Vision-LLMs}}\\
\midrule
InstructBLIP \cite{dai2023instructblip} & 17.50 \\
InstructBLIP CoT & 31.50\\
LLaVA \cite{liu2023llava} & 25.00  \\
LLaVA CoT & 33.50 \\
\midrule
\midrule
\multicolumn{2}{c}{\textbf{(b) The Effect of Two-Turns prompt}}\\
\midrule
GPT-4V (1-turn)& 69.25 \\
GPT-4V CoT (1-turn)& 75.25 \\
GPT-4V Desp + GPT-4 QA (2-turns)& 72.25 \\
GPT-4V Desp + GPT-4 CoT (2-turns)& 78.75 \\
GPT-4V Desp + GPT-4V QA (2-turns)& 79.50\\
GPT-4V Desp + GPT-4V CoT (2-turns)& 80.00\\
\bottomrule
\end{tabular}
\caption{\textbf{Model Performance Comparison.} Section (a) presents performance comparisons of vision large language models, specifically InstructBLIP and LLaVA, with and without the utilization of Chain-of-Thought (CoT) prompting. Section (b) examines the effects of employing two-turn prompts on GPT-4V performance. }
\label{table:ablation}
\end{table}

\subsection{The Effect of Two-turns Prompt}
While examining the output of ``GPT-4V CoT", we observed instances where correct descriptions were followed by incorrect answers. To simplify this generation process, we conducted experiments to divide the recognition and reasoning into two turns, on the \emph{text} setting, as illustrated in Figure~\ref{fig: method} (b). 

To examine the quality of the first turn, i.e., the 
question-relevant
image description, we conducted ablation studies employing GPT-4 in the second turn, which generates responses without image access. 
In the following, we present the prompts we used for different settings of the second turn.
 
\noindent\textbf{GPT-4 QA}: {[`image description']} Based on this image description, does this image depict (A) {[`caption-0']}, or (B) {[`caption-1']}? Note, you must choose one of the two options.

\noindent\textbf{GPT-4 CoT}: {[`image description']} Based on this image description, does this image depict (A) {[`caption-0']}, or (B) {[`caption-1']}? First, analyze the two options, then provide your answer. Note, you must choose one of the two options.

\noindent\textbf{GPT-4V QA}: {[`image description']} Does this image depict (A) {[`caption-0']}, or (B) {[`caption-1']}? Note, you must choose one of the two options.

\noindent\textbf{GPT-4V CoT}: {[`image description']} Does this image depict (A) {[`caption-0']}, or (B) {[`caption-1']}? First, analyze the two options, then provide your answer. Note, you must choose one of the two options.

The results are summarized in Table~\ref{table:ablation} (b). Our observations are threefold: 1) The two-turn prompt notably improves results, from 75.25\% to 79.50\%. 2) The QA performance of GPT-4 reflects the quality of the image descriptions generated by GPT-4V. The ``GPT-4V Desp + GPT-4 CoT (2-turns)" experiment achieved a 78.75\% performance rate, validating the preciseness of GPT-4V's image descriptions. 3) GPT-4V performs slightly better than GPT-4, indicating that while the text format makes reasoning easier, fully enumerating all related information presented in an image remains challenging.

\begin{table}[t]
\small
\centering
\begin{tabular}{l p{1.5cm} r}
\toprule
\textbf{Tag} & \textbf{Correct|Tagged} & \textbf{Accuracy} \\
\midrule
Symbolic & [36 | 41] &  (96.46\%) 87.80\%\\
Series & [21 | 31] &  (96.65\%) 67.74\%\\
Pragmatics & [17 | 24] & (58.82\%) 70.83\% \\
\midrule
Adjective-Color & [40 | 47] & 85.11\% \\
Adjective-Size/Amount & [13 | 24] & 54.17\% \\
Adjective-Animate & [9 | 9] & 100.00\% \\
Adjective-Texture & [8 | 8] & 100.00\% \\
Adjective-Height & [7 | 7] & 100.00\% \\
Adjective-Shape & [6 | 6] & 100.00\% \\
Adjective-Temperature & [5 | 6] & 83.33\% \\
Adjective-Weight & [0 | 3] & 0.00\% \\
Adjective-Age & [2 | 2] & 100.00\% \\
\midrule
Determiner-Numeral & [23 | 27] & 85.19\% \\
\midrule
Object-Centric-Spatial & [9 | 16] & 56.25\% \\
\midrule
Temporal Dynamics & [7 | 16] & 43.75\% \\
\bottomrule
\end{tabular}
\caption{\textbf{Error Analysis by Tag Category.} The table presents the number of correct items versus the total items tagged in each category, alongside the corresponding accuracy. The results reported here are from the experiment ``GPT-4V Desp + GPT-4V CoT'', with an overall accuracy of $80\%$ on text score. The percentages in parentheses present the MTurk Human performance.}
\label{table: error}
\end{table}

\begin{figure*}[t]
\centering
\includegraphics[width=0.95\linewidth]{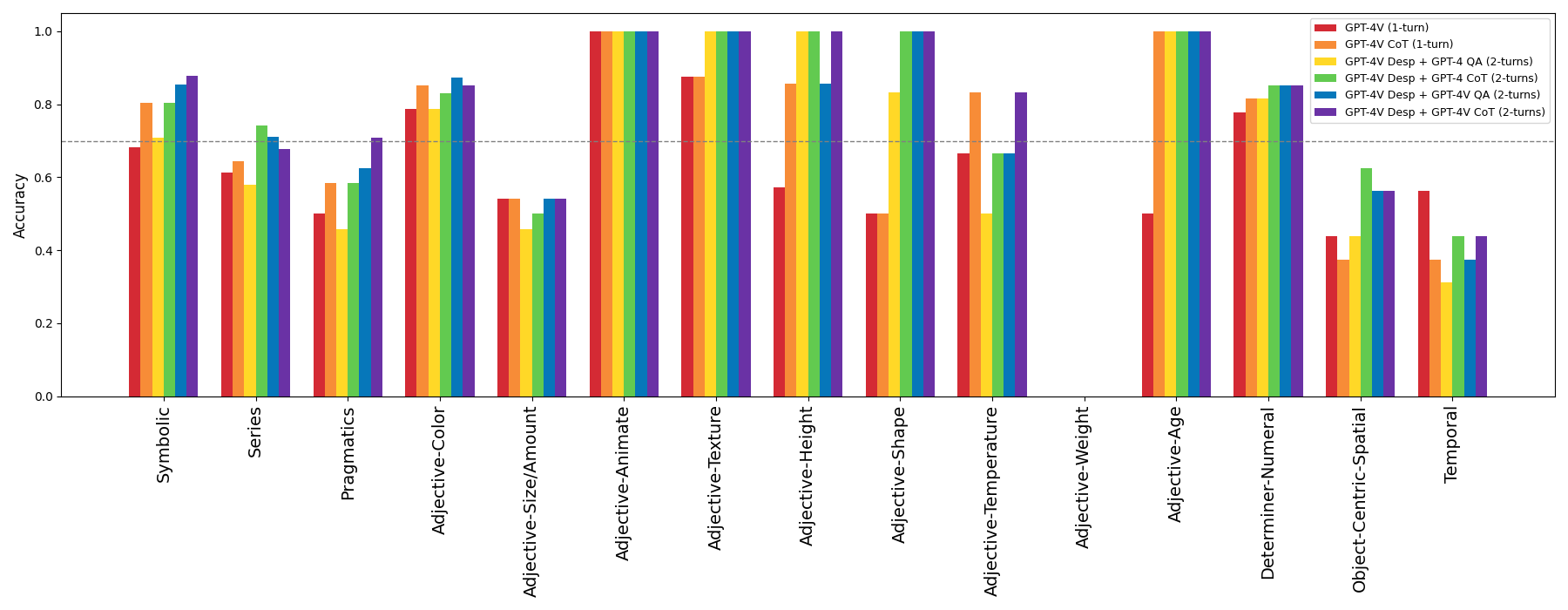}
\caption{\textbf{Error analysis by tag category across different GPT-4V prompt configurations.} Each bar represents the accuracy of a specific experiment configuration on a given tag category.
From left to right, the experiments are: GPT-4V (1-turn), GPT-4V CoT (1-turn), GPT-4V Desp + GPT-4 QA (2-turns), GPT-4V Desp + GPT-4 CoT (2-turns), GPT-4V Desp + GPT-4V QA (2-turns), and GPT-4V Desp + GPT-4V CoT (2-turns).}
\label{fig: error_analysis}
\end{figure*}

\section{Error Analysis}
\vspace{-6pt}
The performance of GPT-4V on the Winoground is remarkable, yet it's essential to pinpoint its limitations for a comprehensive understanding. Our error analysis, which adapts Winoground's tagging categorization, examines GPT-4V's performance on vision-language understanding tasks that require various recognition and reasoning skills. 

The experiment selected for this analysis is "GPT-4V Desp + GPT-4V CoT (2-turns)," which achieves 80\% text score across the dataset. As per Table~\ref{table: error}, GPT-4V is adept at interpreting symbolic categories, which represent symbolic representations, e.g., a child's drawing. 
The "Series" tag indicates that a pair of images comes from the same photographic series, which might include identical individuals and scenes. While humans quickly understand semantic differences, visual similarities can pose challenges to the model, leading to lower accuracy compared to the baseline. The "Pragmatics" tag is for images that require non-literal interpretation, such as understanding idiomatic language or clarifying syntactic ambiguities in captions. Challenges such as discerning metaphors or the specific semantic relationships in prepositional phrases are reflected in the lower accuracy rates for both humans and models compared to the baseline.


In differentiating attributes, objective ones based on visual facts, such as color and shape, are easier to discern. In contrast, more abstract attributes, like size or amount, which require a reference, or weight, which necessitates additional knowledge, are more challenging to identify.

The "Determiner Numeral" category, which usually involves counting, is handled competently. Object-centric spatial tasks require identifying the correct reference frame and understanding the spatial relationships from an object's perspective in the image, which demands an interpretation of the 3D world from a 2D representation; this remains challenging and is generally more difficult than discerning camera-view spatial relationships. Temporal dynamics, which involve inferring past, present, and future events, are still difficult for the model.

\begin{figure*}[t]
\centering
\includegraphics[width=\linewidth]{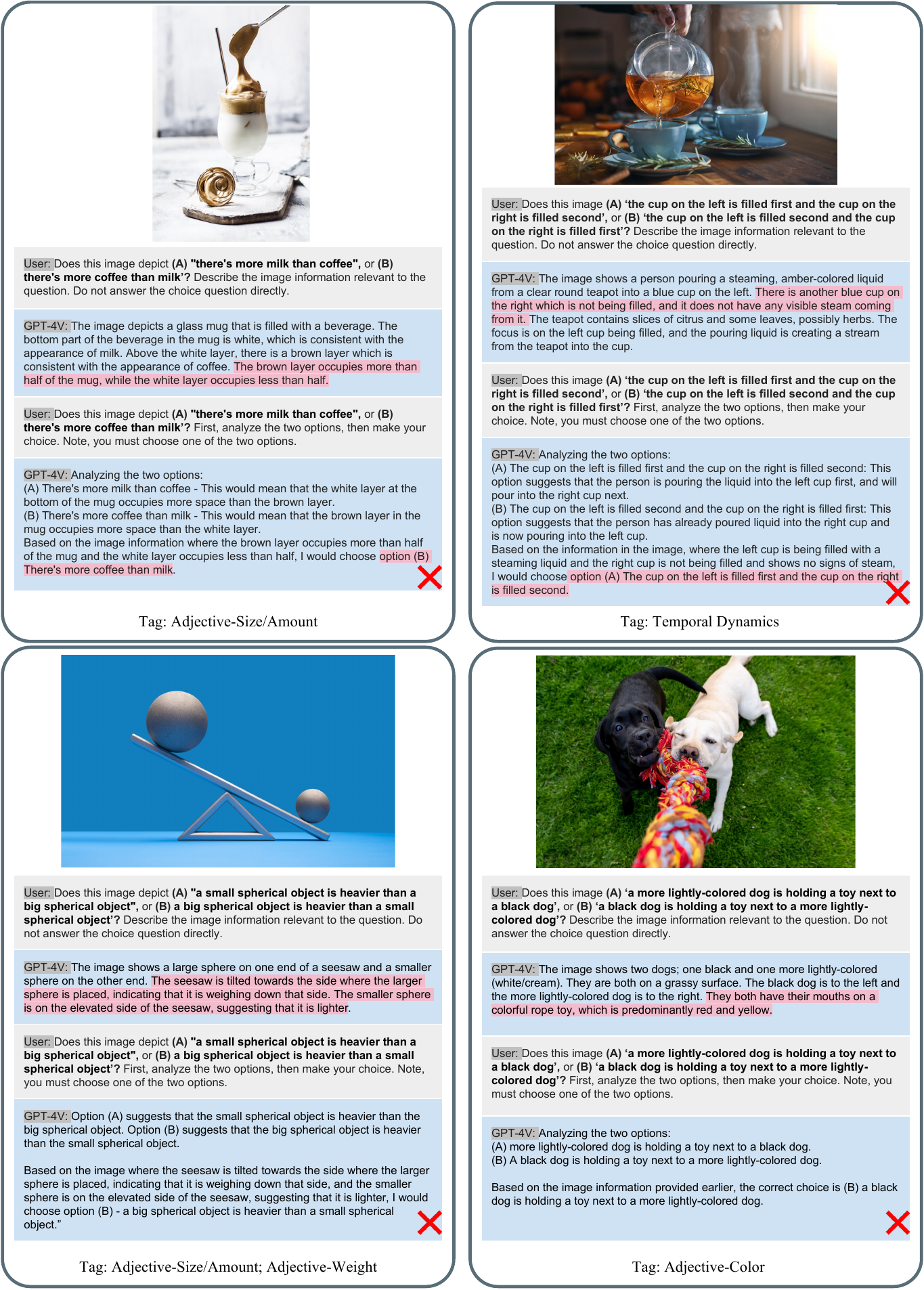}
\caption{Examples of error cases with experiment ``GPT-4V Desp + GPT-4V CoT (2-turns)'', along with the category tags. All images shown here are from Winoground \cite{thrush2022winoground}.}
\label{fig: error1}
\end{figure*}

We present the error analysis in Figure~\ref{fig: error_analysis} for all experimental settings shown in Table~\ref{table:ablation}. The results consistently show that the categories tagged with 'Series, Pragmatics, Size/Amount, Weight, Object-Centric Spatial, Temporal' are comparatively more difficult. Error examples are presented in Figure~\ref{fig: error1}.

\section{Conclusion}
\vspace{-6pt}
In this study, we introduce a ``description then decision" strategy for vision-language tasks. From a neuroscience perspective, humans conduct recognition and reasoning in distinct modules and through multiple steps. From a model training perspective, large language models are proficiently trained on linguistic tasks, and vision encoders have increasingly been aligned with these language models through image captioning. Given a vision-language task, the ``description then decision" approach transforms the task into two well-trained tasks. Although straightforward, our prompt strategy has demonstrated consistent improvements across various models, paving the way for future research into reasoning paradigms for vision-language tasks.

\clearpage
\bibliography{main}
\bibliographystyle{acl_natbib}

\end{document}